\documentclass[runningheads]{llncs}
\usepackage{graphicx}

\usepackage{tikz}
\usepackage{comment}
\usepackage{amsmath,amssymb} 
\usepackage{color}
\usepackage[hidelinks]{hyperref}

\usepackage{amsthm}
\usepackage{booktabs}
\usepackage{algorithm}
\usepackage{algorithmic}
\usepackage{enumitem}
\usepackage[]{subcaption}
\usepackage{makecell}
\usepackage[page]{appendix}
\makeatletter
\newcommand{\@chapapp}{\relax}
\makeatother

\begin{document}
\pagestyle{headings}
\mainmatter
\def\ECCVSubNumber{4882}  

\title{It's All in the Head: \\ Representation Knowledge Distillation through Classifier Sharing}

\titlerunning{Representation Knowledge Distillation through Classifier Sharing}
%
\author{Emanuel Ben-Baruch\thanks{Equal contribution}
\and
Matan Karklinsky$^*$\and
Yossi Biton\and
Avi Ben-Cohen\and \\
Hussam Lawen \and
Nadav Zamir}
\authorrunning{E. Ben-Baruch et al.}
%
\institute{DAMO Academy, Alibaba Group\\
\email{\{emanuel.benbaruch, matan.karklinsky, yossi.biton, avi.bencohen, hussam.lawen, nadav.zamir\}@alibaba-inc.com}}

\maketitle
\begin{abstract}
\label{abstract}

Representation knowledge distillation aims at transferring rich information from one model to another.
Common approaches for representation distillation mainly focus on the direct minimization of distance metrics between the models' embedding vectors.
Such direct methods may be limited in transferring high-order dependencies embedded in the representation vectors, or in handling the capacity gap between the teacher and student models.
Moreover, in standard knowledge distillation, the teacher is trained without awareness of the student's characteristics and capacity.
In this paper, we explore two mechanisms for enhancing representation distillation using classifier sharing between the teacher and student. 
We first investigate a simple scheme where the teacher's classifier is connected to the student backbone, acting as an additional classification head.
Then, we propose a student-aware mechanism that asks to tailor the teacher model to a student with limited capacity by training the teacher with a temporary student's head.
We analyze and compare these two mechanisms and show their effectiveness on various datasets and tasks, including image classification, fine-grained classification, and face verification. 
In particular, we achieve state-of-the-art results for face verification on the IJB-C dataset for a MobileFaceNet model: TAR@(FAR=1e-5)=93.7\%. Code is available at https://github.com/Alibaba-MIIL/HeadSharingKD.
\end{abstract}

\section{Introduction}
\label{introduction}

\begin{figure*}[t!]
\begin{subfigure}[a]{0.5\textwidth}
  \centering
  \includegraphics[width=0.9\linewidth]{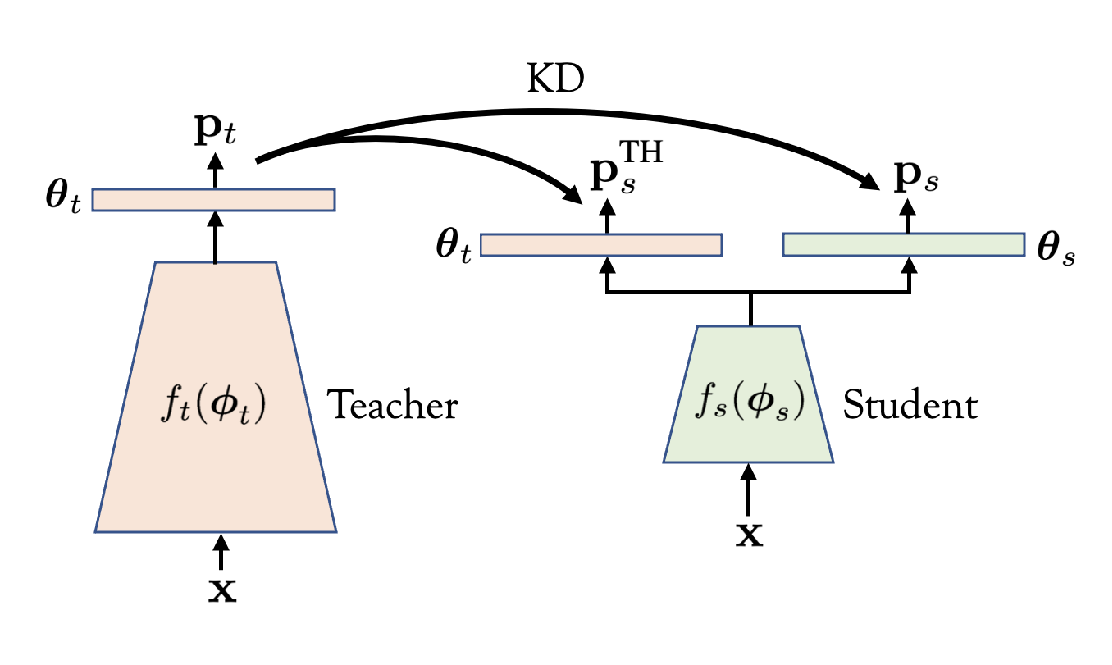}
  \caption{Teacher-Head KD (TH-KD)}
    \label{subfig:approach_TH}
\end{subfigure}%
\begin{subfigure}[h]{0.5\textwidth }
  \centering
  \includegraphics[width=0.9\linewidth]{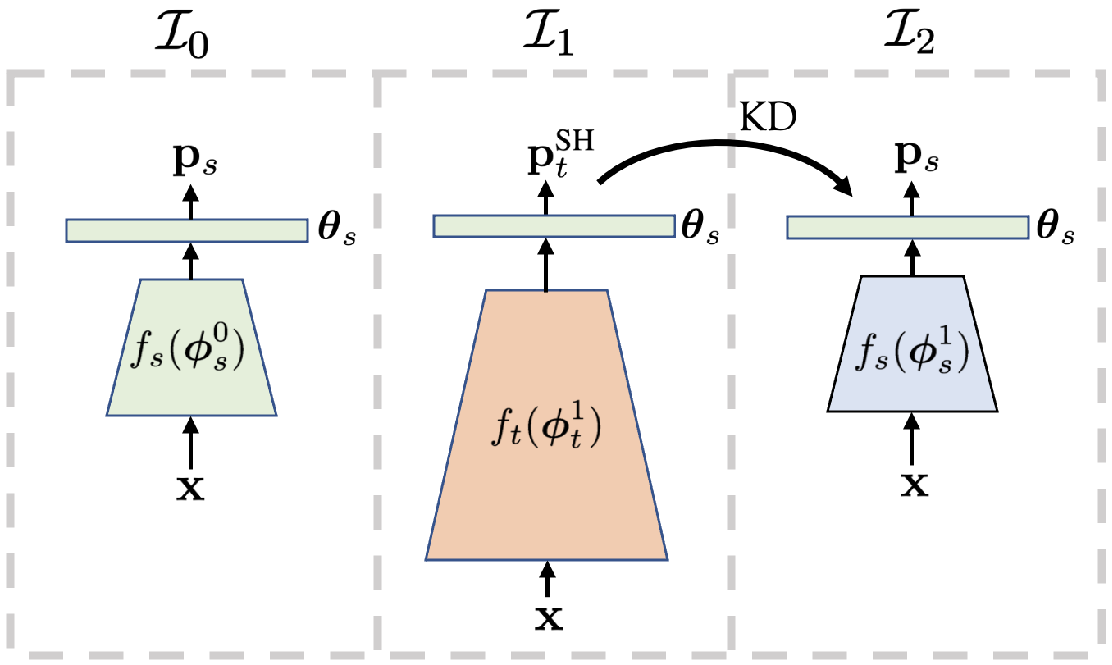}
  \caption{Student-Head KD (SH-KD)}
    \label{subfig:approach_SH}
\end{subfigure}

\caption{\textbf{Illustration of the two proposed schemes based on classifier sharing between teacher and student.} (a) The first approach uses the teacher’s classifier as an auxiliary head, with frozen weights, to aid the student learning. (b) In the second approach, we copy the classifier's weights of a pre-trained student to the teacher's classifier (and freeze them) to regularize the representation learning of the teacher. Then, a new student is trained using knowledge distillation.}
\label{fig:approach}
\vspace{-5mm}
\end{figure*}

Knowledge distillation (KD) is a commonly used technique for improving the accuracy of a compact model using the guidance of a larger teacher model. 
In the original KD approach \cite{hinton2015distilling}, the teacher's knowledge is transferred to the student by minimizing an objective function that operates only on the final output predictions of the models. Thus, the transferred knowledge from the teacher may be partial and limited.

Representation distillation is often favored for transferring richer information and semantic knowledge from the teacher to the student \cite{romero2015fitnets,tian2020contrastive}. Enabling robust representation transfer may be particularly beneficial for tasks that highly rely on the structure of the representation space and its discrimination quality, such as fine-grained classification \cite{touvron2020grafit}, face recognition \cite{Liu_2021_ICCV}, and image retrieval \cite{chen2021deep}.

Yet, most existing approaches for representation distillation are based on minimizing a loss function as a distance between the teacher and student representation vectors.
For example, the L2 norm can be used as a reconstruction loss between the embedding vectors extracted by the teacher and student models \cite{Liu_2021_ICCV}.
\cite{romero2015fitnets} proposed transferring the knowledge via intermediate representations learned by the teacher's hidden layers, and \cite{zagoruyko2017paying} introduced an approach that encourages the student to mimic the attention maps of the teacher.
In \cite{tian2020contrastive}, they proposed a contrastive loss for representation distillation based on maximizing a lower-bound to the mutual information between the teacher and student embedding vectors.

Such direct approaches, which aim at minimizing a distance metric between the embedding vectors, may be limited in transferring the representational knowledge from the teacher to the student \cite{shen2021backwardcompatible}, as the discriminative power may reside in singular dimensions or be hidden in complex correlations between the embedding dimensions.
In addition, as the teacher's complexity can be significantly higher than the student's complexity, the student may not have the capacity to mimic the representation space of the teacher. This is known as the \textit{capacity gap} problem \cite{mirzadeh2019improved}.
Therefore, we ask for learning strategies that will support the representation distillation process by bridging the gap between the teacher and the student.

In particular, we investigate the ability to use the models' classifiers to aid the training process.

A model's classifier captures essential information regarding the representation space structure and the discrimination capabilities of the model. For example, in \cite{kang2020decoupling}, they tackle the imbalance in long-tail recognition by adjusting only the classifier weights.
Previous works used the classifier weights of a pre-trained model when training a new model for backward representation compatibility \cite{shen2021backwardcompatible}, or for unsupervised domain adaptation \cite{liu2021cycle}.
Inspired by these approaches, in this work we propose to enhance representational knowledge distillation by sharing the classifiers between the teacher and the student models.

Specifically, we explore two methods that deploy classifier weights sharing between the teacher and the student. In the first method, the teacher’s classifier is used to constrain the student representation learning by connecting the teacher's classifier to the student backbone as an additional head (with frozen parameters). We name this approach \emph{Teacher-Head KD}, denoted by TH-KD (Fig.~\ref{subfig:approach_TH}). Sharing the classification boundaries of the teacher in the student optimization process may help shape its representation space to be similar to the one of the teacher.
Closest to this scheme is the work presented in \cite{Yang2021KnowledgeDV}, where the student is trained such that the teacher's and the student's embeddings produce the same output when passed through the teacher's classifier. 
Herein, we propose a more generalized scheme. First, we propose to use a combination of both the teacher’s and the student’s classifiers in the distillation loss. Second, during inference, we suggest using the predictions from both classifiers, aggregated by a weighted sum. This way, the TH-KD scheme enables to deploy different training and inference configurations. In particular, it can be configured such that the teacher head completely replaces the student head.

Next, we propose a student-aware mechanism for representation distillation based on sharing a student's classifier with the teacher learning process.
In this approach, we first train a temporary student model. Then, the parameters of the student's classifier are used to initialize the teacher's head and are fixed during the training of the teacher's backbone.
A final student model is trained using the representation distillation loss.
We name this approach \emph{Student-Head KD}, denoted by SH-KD (Fig.~\ref{subfig:approach_SH}).
While a conventional training of a teacher can produce a high-quality feature space in terms of class separation and test accuracy results, in practice, it may be hard for the student to follow the complexity of the teacher's representation due to the limited capacity of the student backbone. 
Training the teacher with the student's head enforces the teacher's backbone to learn features that better suit the capacity of the student.
This way, the student can mimic the teacher's representation more easily.

Methods for training a student-aware teacher for knowledge distillation were proposed in \cite{zhou2021meta} and \cite{park2022learning}. A meta-learning framework was introduced in \cite{zhou2021meta}, that enables to train a teacher with the feedback from the distilled student performance.  In \cite{park2022learning}, they suggested to train a teacher along with the student branches jointly to obtain student-friendly representations.
In this paper, we propose a simple approach that does not require sophisticated mechanisms such as bi-level optimization or joint training of teacher and student but focuses on sharing the student's classifier with the teacher.

We analyze the capabilities of the two explored schemes and compare them to other KD methods. In particular, by measuring the angle between the features of the teacher and the student's, we show that the student-aware mechanism of SH-KD enables the student to learn features that are closer to the ones of the teacher compared to other baseline approaches. 
Both TH-KD and SH-KD methods are tested for various tasks on several datasets: CIFAR-100 \cite{krizhevsky2009learning}, Stanford-cars \cite{stanford_cars_2013} FoodX-251 \cite{foodx_dataset_2019} and for face verification on the IJB-C dataset \cite{8411217}. Specifically, using the SH-KD scheme, we achieve state-of-the-art results on the IJB-C dataset  when using MobileFaceNet model: TAR(1e-5)=93.7\%.

The contribution of the paper can be summarized as follows:
\begin{itemize}
    \item We explore and analyze two mechanisms for representation distillation based on classifier sharing between the teacher and the student models: TH-KD and SH-KD. 
    These techniques are easy-to-implement and complementary to other knowledge distillation approaches.
    \item We introduce SH-KD: a novel student-aware mechanism for representation distillation that enables tailoring the teacher model to a specific student, and to mitigate the capacity gap between the teacher and the student, by training the teacher with a temporary student’s head.
    \item Our methods achieve consistent accuracy improvement for various settings, across datasets and on different architectures, including obtaining state-of-the-art results for face verification on the IJB-C dataset.
\end{itemize}

\section{Representational Knowledge Distillation through Classifier Sharing}
In this section, we introduce two approaches based on classifier sharing between the teacher and student models to facilitate the representation distillation process. In the first scheme the teacher's classifier is used to constrain the student representation learning. In the second scheme, a student's classifier is used to regularize the representation learning of the teacher.

\subsection{Problem Formulation}
Given a teacher model $f_t$, we aim at training a smaller student model $f_s$, guided by the teacher. For a given input sample $\mathbf{x}$, we denote by $\mathbf{z}_t = f_t(\mathbf{x}; \boldsymbol{\phi}_t)$ and $\mathbf{z}_s=f_s(\mathbf{x};\boldsymbol{\phi}_s)$ the representation (embedding) vectors of the teacher and student models, respectively, where $\boldsymbol{\phi}_t$ and $\boldsymbol{\phi}_s$ are the teacher and student models' parameters, respectively. The teacher's classifier is defined by $g_t(\mathbf{z})=W_t\mathbf{z}+\mathbf{b}_t$, and the student's classifier is defined by $g_s(\mathbf{z})=W_s\mathbf{z}+\mathbf{b}_s$. The final prediction is given by applying the softmax activation $h(\cdot)$: $\mathbf{p}_t=h(g_t(\mathbf{z}_t))$, and $\mathbf{p}_s=h(g_s(\mathbf{z}_s))$ for the teacher and the student, respectively.
For simplicity, we denote the classifier weights and bias terms of the teacher and the student by $\boldsymbol{\theta}_t = \{W_t, \mathbf{b}_t\}$ and $\boldsymbol{\theta}_s = \{W_s, \mathbf{b}_s\}$, respectively.

For a given training sample $\mathbf{x}$ and a corresponding ground-truth label vector $\mathbf{y}$, a general form of the loss function used to train the student model can be written as:
\begin{equation}\label{base_loss}
    \mathcal{L}=\mathcal{L}_{\text{CE}}(\mathbf{p}_s, \mathbf{y})+\alpha\mathcal{H}(\mathbf{p}_s, \mathbf{p}_t) + \beta\mathcal{D}(\mathbf{z}_s, \mathbf{z}_t),
\end{equation}
where $\mathcal{L}_{\text{CE}}(\cdot)$ is the cross-entropy loss, and $\mathcal{H}(\cdot)$ is the knowledge distillation distance function between the probabilistic outputs of the teacher and student models, e.g.\ the KL-divergence \cite{hinton2015distilling}. The term $\mathcal{D}(\cdot)$ refers to a distance metric applied on the representation vectors of the teacher and student models, as the L2 loss, cosine distance or a contrastive loss \cite{tian2020contrastive}, where
$\alpha$ and $\beta$ are constant hyper-parameters that control the contribution of each loss term.

In particular, the L2 loss for representation distillation was found to be useful for face recognition \cite{Liu_2021_ICCV} and other general fine-grained classification tasks. The L2 loss is computed by the euclidean distance of the normalized embedding vectors. We term this loss as L2E.
Note that in case that the embedding dimensions of the teacher and student differ, we add a linear transformation to the architecture's head to match their dimensions.

While the KD loss $\mathcal{H}(\cdot)$ enables the transfer of valuable knowledge encapsulated in the soft predictions of the teacher, minimizing the representation loss $\mathcal{D}(\cdot)$, enforces the embedding space of the student to be aligned with the teacher's embedding space. Thus, these loss terms are complementary and together they enable a robust knowledge transfer from the teacher to the student.
\vspace{-3mm}
\subsection{Teacher-Head Sharing (TH-KD)}
We aim to utilize the discrimination information represented by the classification decision boundaries of the teacher to guide the student model in the optimization process. In this scheme, we propose to use the teacher's classifier as an auxiliary head for training the student model. Let $\mathbf{p}_s^{\text{TH}}=h(g_t(\mathbf{z}_s))$ be the prediction vector output from the teacher's classifier for a given student's embedding input, we combine the KD losses computed for the two classifiers as follows,
\begin{equation}
    \mathcal{H}' = (1-\alpha^{\text{TH}})\mathcal{H}(\mathbf{p}_s, \mathbf{p}_t) + \alpha^{\text{TH}}\mathcal{H}(\mathbf{p}_s^{\text{TH}}, \mathbf{p}_t),
\end{equation}
where $\alpha^{\text{TH}}$ is a constant hyper-parameter that balances between the losses of the two classification heads. Similarly, the classification loss is given by,
\begin{equation}\label{TH_CE_loss}
    \mathcal{L}_{\text{CE}}' = (1-\alpha^{\text{TH}})\mathcal{L}_{\text{CE}}(\mathbf{p}_s, \mathbf{y}) + \alpha^{\text{TH}}\mathcal{L}_{\text{CE}}(\mathbf{p}_s^{\text{TH}}, \mathbf{y}).
\end{equation}
In inference time, the final prediction can be obtained by combining the head outputs:
\begin{equation}\label{TH_CE_inference}
    \mathbf{p}_s' = (1-\alpha^{\text{TH}})\mathbf{p}_s + \alpha^{\text{TH}}\mathbf{p}_s^{\text{TH}}.
\end{equation}
This method, named TH-KD, is illustrated in Fig.~\ref{subfig:approach_TH}. Note that for $\alpha^{\text{TH}} = 1$, the student's head is simply the teacher's head whose weights are fixed during the training. Setting $\alpha^{\text{TH}} = 0$ leads to the conventional scheme for knowledge distillation.

Incorporating the teacher-head loss encourages the student to mimic the representation space of the teacher while resolving its high dimensional dependencies. In section \ref{sec: analysis} (Fig.~\ref{fig:MSC}), we show that the TH-KD scheme leads to an enhanced representation quality, which is expressed in terms of a higher inter-class separability and a lower intra-class variation of the embedding space. 

\subsection{Student-Head Sharing (SH-KD)}
The second approach aims at tackling the limited capacity of the student in the distillation process. In a conventional knowledge distillation, the teacher model is trained independently and in isolation from the student training process. 
Typically, the capacity of the teacher model is higher than the capacity of the student model, and thus the features learned by the teacher may not be applicable for the student training.

To this end, we propose to train a teacher while considering the limited capacity of the student, by initializing the teacher's classifier with the weights of a temporary student's head and fixing them during the training.
This process can be viewed as a regularization mechanism that enforces the teacher to learn useful features suited for the student's limitations.
\newline
The method can be depicted as a three-step training procedure:

\textbf{step-$\mathcal{I}_0$}: A student model is trained, with KD or without, providing a backbone and a classifier head with parameters, $\{\boldsymbol{\phi}_s^0, \boldsymbol{\theta}_s \}$.

\textbf{step-$\mathcal{I}_1$}: A teacher model is trained by initializing and fixing its classifier with $\boldsymbol{\theta}_s$ to obtain the teacher model parameters, $\{\boldsymbol{\phi}_t, \boldsymbol{\theta}_s \}$.

\textbf{step-$\mathcal{I}_2$}: A student model is trained using the loss in equation (\ref{base_loss}), with the teacher model obtained in step $\mathcal{I}_1$, to produce the final student parameters:  $\{\boldsymbol{\phi}_s^1, \boldsymbol{\theta}_s \}$.

The approach, named SH-KD, is illustrated in Fig.~\ref{subfig:approach_SH}. Note that for cases where the embedding dimensions of the teacher and the student are not the same, we add a linear transformation to the model's backbone.
SH-KD method offers an easy-to-implement yet effective scheme that enables to tailor the teacher model to comply with the student's capacity during the teacher's training at the cost of an additional training iteration.

Indeed, the accuracy of the teacher may be decreased when using the SH-KD scheme compared to the accuracy obtained by a teacher that was trained conventionally. 
However, in knowledge distillation, we do not aim at optimizing the teacher, but rather to improve the accuracy of the student. Specifically, a lower accuracy of the teacher does not necessarily lead to a lower accuracy of the student.
The same observation was made in the teacher assistant (TA) technique \cite{mirzadeh2019improved}. Instead of using a teacher model with the largest capacity which produces the highest accuracy, it was shown that under some conditions, a teacher model with intermediate-size (teacher assistant) can provide superior performance for the student. While both TA and SH-KD offer a way to mitigate the capacity gap between the teacher and the student, SH-KD enables to tailor the teacher to the specific student at hand.

In section \ref{sec: analysis}, we show that training with SH-KD leads to a higher similarity between the student and teacher representations, accompanied by an improvement in student accuracy. 

\begin{figure} [t!]
  \centering
  \includegraphics[scale=0.23]{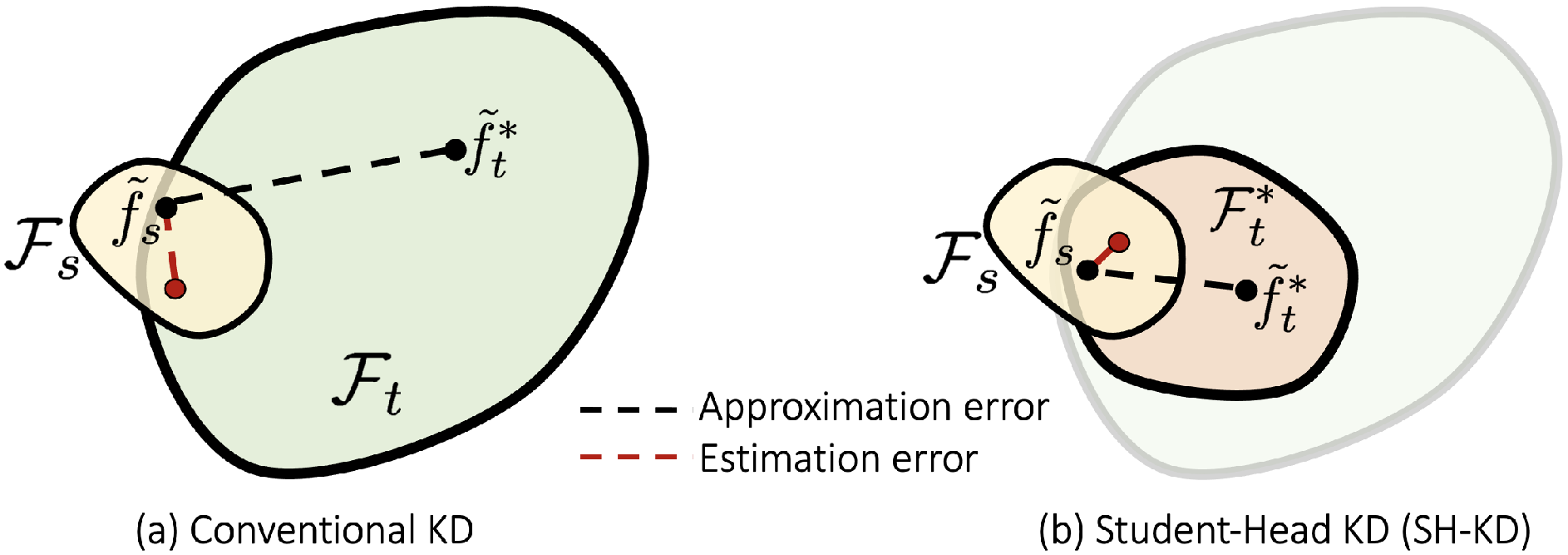}
  \caption{\textbf{A conceptual comparison between baseline KD and the proposed SH-KD scheme.} 
  Training with the SH-KD scheme reduces the teacher's capacity and narrows the space of the teacher function class.  In SH-KD, the teacher is aware of the student's capacity and may produce smaller approximation and estimation errors. 
  }
  \label{fig:conceptual_illustration}
  \vspace{-0.25cm}
\end{figure}

\subsubsection{Theoretical Formulation for SH-KD.}
    We follow the works in \cite{lopezpaz2016unifying} and \cite{mirzadeh2019improved} to shed some light on why the SH-KD scheme can be effective for knowledge distillation.
For simplicity, we assume a pure KD loss, i.e. $\mathcal{L} = \mathcal{H}(\mathbf{p}_s, \mathbf{p}_t)$. 
Also, in the following formulation, the model functions include both the backbone and the classifier: the student function is denoted by $\tilde{f}_s = h(f_s \circ g_s)$, and the teacher function is denoted by $\tilde{f}_t = h(f_t \circ g_t)$.
In the case of a baseline knowledge distillation, the expected error of the student model $R(\tilde{f}_s)$ according to the  VC theory \cite{Vapnik1998} can be expressed as,
\vspace{-8pt}
\begin{equation}\label{error_KD}
    R(\tilde{f}_s) - R(\tilde{f}_t) \leq O\bigg(\frac{|\mathcal{F}_s|_C}{n^{r_{st}}}\bigg) + \epsilon_{st},
    \vspace{-6pt}
\end{equation}
where $|\cdot|_C$ is a function class capacity measure, and $\mathcal{F}_s$ is the student function class.
Here, $O(\cdot)$ and $\epsilon_{st}$ are the estimation and the approximation errors of the student, respectively, and $n$ is the number of the training samples. Also, $\frac{1}{2} \leq r_{st} \leq 1$ is the rate of learning which relates to the training difficulty. A difficult task is characterized by a smaller $r_{st}$ while for an easy task, $r_{st}$ is close to $1$. Similarly, the expected error for a student model, learned by the SH-KD scheme is given by,
\begin{equation}\label{error_SH}
    \vspace{-3pt}
    R(\tilde{f}_s) - R(\tilde{f}_t^*) \leq O\bigg(\frac{|\mathcal{F}_s|_C}{n^{r_{st}^*}}\bigg) + \epsilon_{st}^*,
    \vspace{-3pt}
\end{equation}
where $\tilde{f}_t^{*}$ is the function of the teacher model trained with the student's head: $\tilde{f}_t^* = h(f_t \circ g_s)$. Here, $r_{st}^*$ and $\epsilon_{st}^*$ are the learning rate, and the approximation error of the student learned using the SH-KD teacher. Under the worst-case assumption, the classification error of the SH-KD teacher is higher than the conventional teacher's, i.e. $R(\tilde{f}_t^*) = R(\tilde{f}_t) + \delta$, for $\delta \geq 0$. As aforementioned in the previous section, our objective is to minimize $R(\tilde{f}_s)$. A lower teacher error $R(\tilde{f}_t)$ does not necessarily lead to a lower $R(\tilde{f}_s)$. 
Consequently, we can write the upper bound for $R(\tilde{f}_s) - R(\tilde{f}_t)$, as follows:
\begin{align}
        R(\tilde{f}_s) - R(\tilde{f}_t) & = R(\tilde{f}_s) - R(\tilde{f}_t^*) + \delta \\
        & \leq O\bigg(\frac{|\mathcal{F}_s|_C}{n^{r_{st}^*}}\bigg) + \epsilon_{st}^* + \delta.
\end{align}
Therefore, in order for SH-KD to outperform the baseline KD, the following equation should be satisfied:
\begin{align}\label{eq:theoretic_condition}
        O\bigg(\frac{|\mathcal{F}_s|_C}{n^{r_{st}^*}}\bigg) + \epsilon_{st}^* + \delta \leq O\bigg(\frac{|\mathcal{F}_s|_C}{n^{r_{st}}}\bigg) + \epsilon_{st}.
\end{align}
The task of learning from a teacher that was trained with a student classifier is assumably simpler compared to learning from a conventional teacher because the teacher trained in the SH-KD scheme is tailored to the student's capacity. Thus, typically $r_{st}^* \geq r_{st}$. This is supported experimentally in section \ref{sec:convergence} by comparing the convergence rate of each training scheme as shown in Fig.~\ref{fig:accelerate}.

\begin{figure*}[t!]
\centering
\begin{subfigure}[a]{.33\textwidth}
  \centering
  \includegraphics[width=1 \linewidth]{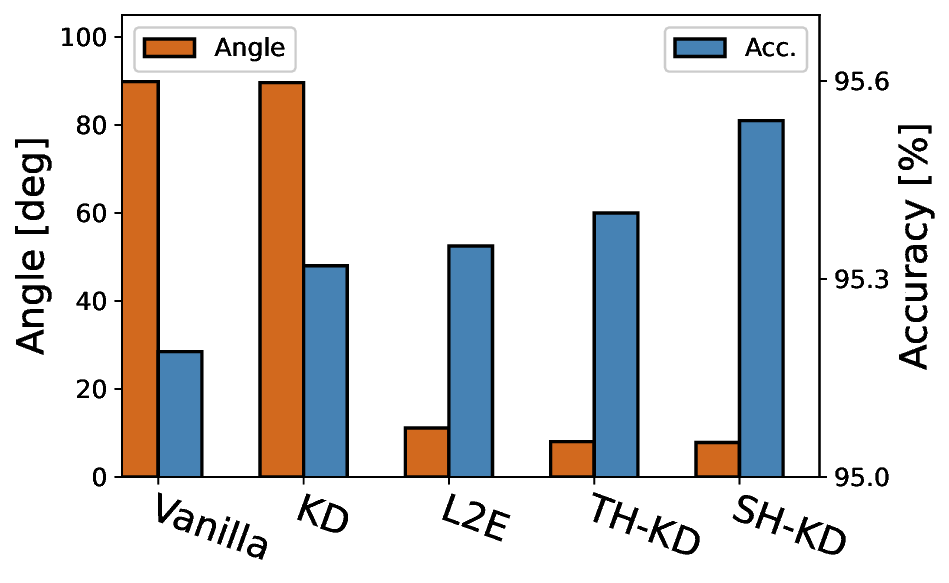}
  \caption{Stanford-cars dataset}
\end{subfigure}%
\begin{subfigure}[h]{.33\textwidth }
  \centering
  \includegraphics[width=1 \linewidth]{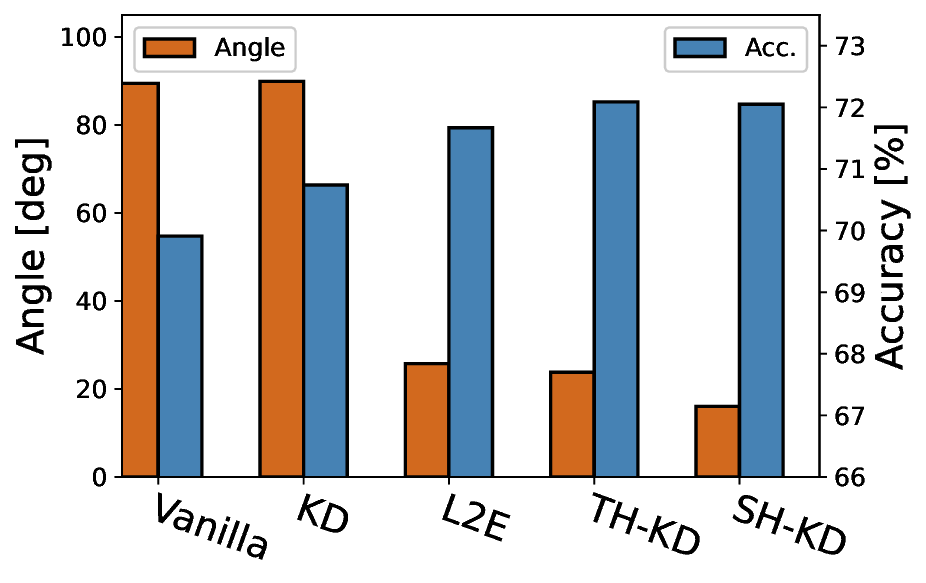}
  \caption{FoodX-251 dataset}
\end{subfigure}
\begin{subfigure}[h]{.33\textwidth }
  \centering
  \includegraphics[width=1\linewidth]{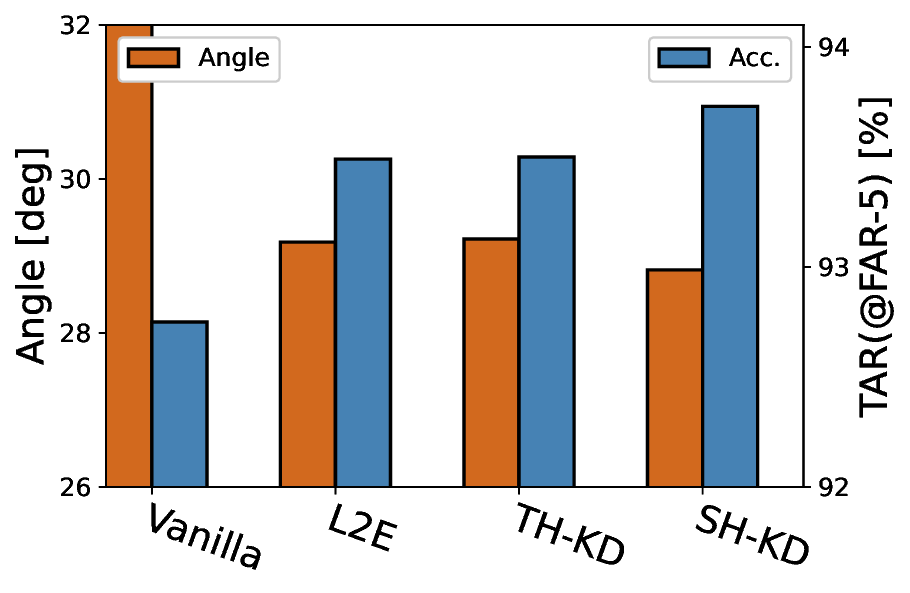}
  \caption{IJB-C dataset}
\end{subfigure}
\caption{\textbf{Average angle between the teacher and student embedding vectors, and student model accuracy.} Both TH-KD, and to a greater extent SH-KD, reduce the angle between the teacher and student embedding vectors and improve test accuracy.}

\label{fig:angle}
\vspace{-3mm}
\end{figure*}

Equation (\ref{eq:theoretic_condition}) is also reasonable under the assumption that $\epsilon_{st}^* \leq \epsilon_{st} - \delta$; while the SH-KD teacher may have lower accuracy than the baseline teacher, as expressed in a positive expected error gap $\delta$, the fact that in SH-KD the teacher and the student share the same classifier encourages a smaller approximation error.
This is supported by Fig.~\ref{fig:angle}; student and teacher features are closer in SH-KD than in baseline KD. In other words, to obtain better performance with SH-KD, the lower approximation error $\epsilon_{st} - \epsilon_{st}^*$ should compensate for the teacher's accuracy drop $\delta$.
A conceptual comparison between the SH-KD training scheme and the baseline KD approach is illustrated in Fig.~\ref{fig:conceptual_illustration}. In case the drop of the teacher accuracy is too high,  equation (\ref{eq:theoretic_condition}) may not hold. Another failure case is when the classifier of the temporary student consists of a deficient decision boundary which can limit the learning ability of the teacher, i.e. $r_{st}^*$ may be small and close to 1/2 . Note that as in \cite{lopezpaz2016unifying} and \cite{mirzadeh2019improved}, the inequality (\ref{eq:theoretic_condition}) holds in the asymptotic regime, and is based on loose upper bounds. Yet, it offers motivation for using SH-KD, and highlights its potential advantages and failure cases.

\section{Experiments}
\label{results}

\label{table_fine_grained}

\begin{table}[t!]

\setlength{\tabcolsep}{4.5pt}
\begin{center}
\begin{tabular}{|l||c|c|c|c|}
\hline
\thead{\textbf{Teacher} \\ \textbf{Student}} & \thead{\textbf{ResNet32x4}\\\textbf{ResNet8x4}}  & \thead{\textbf{ResNet110}\\\textbf{ResNet20}} & \thead{\textbf{ResNet110}\\\textbf{ResNet32}} & \thead{\textbf{ResNet56}\\\textbf{ResNet20}} \\
\hline
Teacher & 79.42 & 74.31 & 74.31 & 72.34 \\
Student & 72.50  & 69.06 & 71.14 & 69.06 \\
\hline

KD \cite{hinton2015distilling} & 73.33  & 70.67 & 73.08 & 70.66 \\
FitNets \cite{romero2015fitnets} & 73.50  & 68.99 & 71.06 & 69.21  \\
AT \cite{passalis2019learning} & 73.44  & 70.22 & 72.31 & 70.55 \\
PKT \cite{passalis2019learning}  & 73.64 & 70.25 & 72.61  & 70.34 \\
AB \cite{heo2018knowledge} & 73.17 & 69.53 & 70.98  & 69.47  \\
FT \cite{kim2020paraphrasing} & 72.86 & 70.22 & 72.37 & 69.84 \\
FSP \cite{8100237} & 72.62 & 70.11 & 71.89 & 69.95 \\
NST \cite{huang2017like} & 73.30 & 69.53 & 71.96 & 69.60  \\
CRD \cite{tian2020contrastive}  & 75.51 & 71.46 & 73.48 & 71.16 \\
CRD+KD \cite{tian2020contrastive} & 75.46  & 71.56 & 73.75 & \textbf{71.63}  \\
SRRL \cite{Yang2021KnowledgeDV} & 75.92 & 71.51 & 73.80 & 71.44 \\
\hline
TH-KD & 75.24   & 71.61 & 73.32 & 71.58 \\
SH-KD & \textbf{75.94}  & \textbf{71.65} & \textbf{73.77} & 71.36  \\
\hline
TH-KD + CRD  & 75.59 & 71.92 & 74.13 & \textbf{72.22} \\
SH-KD + CRD & \textbf{76.61} & \textbf{72.12} & \textbf{74.16} & 72.05 \\
\hline
\end{tabular}
\vspace{0.2cm}
\caption{\textbf{Test accuracy (\%) on CIFAR100 dataset.} We follow the same protocol as in the CRD work \protect\cite{tian2020contrastive}. SH-KD achieves superior results. When combined with the CRD loss, further improvement is obtained in most experiments.}
\label{tbl:cifar100_same}
\end{center}
\vspace{-20pt}
\end{table}

In this section, we report our main results on three domains: image classification, fine-grained classification and face-verification. 
Also, we study the impact of the proposed schemes, TH-KD and SH-KD, on the representation distillation quality.
The training details are provided in the Appendix.
\subsection{Benchmark Results}
\subsubsection{CIFAR-100}

The CIFAR-100 \cite{krizhevsky2009learning} consists of 50K training images (500 samples per class) and 10K test images. In 
Table \ref{tbl:cifar100_same} we show the results obtained by our two explored schemes, TH-KD and SH-KD, and compare them to previous approaches for KD. 
As the proposed schemes are complementary to other methods with different objectives, we tested the TH-KD and SH-KD with the original KD loss and with CRD \cite{tian2020contrastive}. When combined with CRD, both TH-KD and, to a greater extent SH-KD, outperform all the baseline approaches.

\subsubsection{Fine-grained Classification}
We evaluated our methods on two fine-grained classification datasets, Stanford-cars \cite{stanford_cars_2013} and FoodX-251 \cite{foodx_dataset_2019}.
Stanford-cars contains 196 classes and consists of 8,041 training images and 8,144 test images. FoodX-251 \cite{foodx_dataset_2019} contains 251 classes and consists of 118K training images and 28K test images.

We tested six training configurations with different architectures for the teacher and student models. As teachers, we used TResNet-M and TResNet-L \cite{ridnik2020TResNet}, and ResNet101 \cite{resnet_2015}.
As students, we used once-for-all (OFA) models \cite{ofa_2020}; OFA-595 and OFA-62, and small ResNet variants; ResNet18 and ResNet26.
The OFA architectures were designed for cost-effective mobile deployment.

We compared five training regimes: a vanilla training without any KD loss, regular KD, training with L2E for representation distillation, and the proposed approaches TH-KD and SH-KD.

We summarize the results obtained on Stanford-cars and FoodX-251 in Table \ref{tbl:stanford_cars} and Table \ref{tbl:foodx}, respectively. 
On the Stanford-cars dataset, the SH-KD method was consistently superior in all training configurations. Interestingly, for Stanford-cars, regular KD degrades the accuracy compared to a vanilla training. On the FoodX-251 dataset, the highest results were achieved by TH-KD or SH-KD. The TH-KD method was superior in four out of the six tested configurations.

\begin{table*}[t!]
\setlength{\tabcolsep}{3pt}
\begin{center}
\begin{tabular}{|l||c|c|c|c|c|c|}
\hline
%
%
 \thead{Teacher \\Student} & \thead{TResNetM\\OFA-62} & \thead{TResNetM\\OFA-595} & \thead{TResNetL\\OFA-62} & \thead{TResNetL\\OFA-595} & \thead{ResNet101\\ResNet18} & \thead{ResNet101\\ResNet26} \\
\hline
Teacher & 95.53 &  95.53 & 96.19 & 96.19 & 95.63 & 95.63 \\
\hline
Vanilla & 94.66 & 95.41 & 94.66 & 95.41 & 94.66 & 95.20 \\
KD & 94.51 & 95.36 & 94.62 & 95.31 & 94.29 & 94.95 \\
L2E & 95.11 & 95.37 & 94.94 & 95.27 & 94.73 & 95.27 \\
TH-KD & 95.16 & 95.28 & 95.08 & 95.38 & 94.83 & 95.19 \\
SH-KD & \textbf{95.21}& \textbf{95.52} & \textbf{95.13} & \textbf{95.46} & \textbf{94.98} & \textbf{95.38} \\
\hline
\end{tabular}
\vspace{1pt}
\caption{\textbf{Test \emph{accuracy} (\%) on Stanford-cars.} The SH-KD method outperforms other approaches  consistently in all training configurations.}
\label{tbl:stanford_cars}
\end{center}
\vspace{-20pt}
\end{table*}

\begin{table*}[t!]
\setlength{\tabcolsep}{3pt}
\begin{center}
\begin{tabular}{|l||c|c|c|c|c|c|}
\hline
 \thead{Teacher \\Student} & \thead{TResNetM\\OFA-62} & \thead{TResNetM\\OFA-595} & \thead{TResNetL\\OFA-62} & \thead{TResNetL\\OFA-595} & \thead{ResNet101\\ResNet18} & \thead{ResNet101\\ResNet26} \\
\hline
Teacher & 76.36 & 76.36 & 77.11 & 77.11 & 75.51 & 75.51 \\
\hline
Vanilla & 69.91 & 73.80 & 69.91 & 73.80 & 67.08 & 70.80 \\
KD & 70.87 & 73.98 & 70.74 & 74.04 & 67.88 & 71.75 \\
L2E & 71.62 & 74.62 & 71.67 & 74.65 & 69.19 & 72.99 \\
TH-KD & \textbf{71.99} & \textbf{75.23} & 72.09 & \textbf{75.20} & \textbf{69.59} & 72.79 \\
SH-KD & 71.94 & 74.56 & \textbf{72.29} & 74.75 & 69.17 & \textbf{73.12} \\
\hline
\end{tabular}
\vspace{1pt}
\caption{\textbf{Test \emph{accuracy} (\%) on FoodX-251.} The highest results are achieved by the TH-KD or SH-KD approaches. TH-KD is superior in most of the training configurations.}
\label{tbl:foodx}
\end{center}
\vspace{-20pt}
\end{table*}





\begin{table*}[t!]
\setlength{\tabcolsep}{4.5pt}
\begin{center}
\begin{tabular}{|l||l||c|c|c|}
\hline
\thead{\textbf{Method}} & \thead{\textbf{Model}} & \thead{\textbf{TAR@}\\\textbf{FAR=1e-6}} & \thead{\textbf{TAR@}\\\textbf{FAR=1e-5}} &
\thead{\textbf{TAR@}\\\textbf{FAR=1e-4}}\\
\hline
Martinez* & MobileFaceNet & -- & 92.20 & 94.70  \\
L2E+ES-sampling* & MobileFaceNet &-- & 93.20 & 95.39  \\
L2E+IS-sampling* & MobileFaceNet &  -- & 93.25 & 95.49  \\
\hline
Teacher, Vanilla& R100  & 91.49 &  95.52 & 97.00  \\
Teacher, SH-KD & R100  & 90.88 &  95.58 & 97.03  \\
\hline
Vanilla & MobileFaceNet & 88.72 & 92.75 & 95.42  \\
L2E & MobileFaceNet & 88.47  & 93.49 & 95.48  \\
TH-KD (Ours) & MobileFaceNet & 89.82 & 93.50 & 95.48  \\
SH-KD (Ours) & MobileFaceNet & \textbf{90.24}& \textbf{93.73} & \textbf{95.64} \\
\hline
\end{tabular}
\vspace{1pt}
\caption{\textbf{Results on the IJB-C dataset.} The reported results of the first three rows, denoted by *, were taken from the papers \protect\cite{martinez2021benchmarking} and \protect\cite{Liu_2021_ICCV}.
}
\label{tbl:face}
\end{center}
\vspace{-35pt}
\end{table*}

\subsubsection{Face Verification}
We evaluated our methods on the face verification task of the IJB-C dataset \cite{8411217}. 
For training, we used a refined version of the popular MS-Celeb-1M dataset \cite{guo2016msceleb1m} named MS1MV3 \cite{deng2019lightweight} which contains about 93K identities and 5.2M images.
We used a ResNet-like network \cite{resnet_2015}, R100 as a teacher, and MobileFaceNet as a student \cite{chen2018mobilefacenets}. 
We used the L2 loss between the embedding features of the teacher and the student (L2E) as the representation distillation loss, and the large-margin cosine loss, CosFace \cite{Wang_2018_CVPR} as the base loss.

In Table 4 we show the results obtained by our approaches, TH-KD and SH-KD, and compare them to other baselines and previous state-of-the-art methods \cite{Liu_2021_ICCV}. 
We report three common metrics for evaluating the performance of the IJB-C dataset: TAR(@FAR=1e-6), TAR(@FAR=1e-5) and  TAR(@FAR=1e-4). 
In the vanilla training, we used the CosFace loss only. The TH-KD scheme was performed with $\alpha^{\text{TH}}=1$ in equation (\ref{TH_CE_loss}). We also report the results obtained by the teacher models: the regular teacher and the teacher training with the student's classifier following the SH-KD method. As can be seen, both teachers provide similar metric results.

The student model trained using the TH-KD scheme outperforms the baseline approaches considerably for the  TAR(@FAR=1e-6) metric, improving the L2E method from 88.47\% to 89.82\%. Using the SH-KD scheme, we obtain a significant improvement compared to the other baselines and previous state-of-the-art approaches in all the tested metrics. For example, the TAR(@FAR=1e-6) metric is improved to 90.24\%.

\begin{figure}[t!]
\centering
\begin{subfigure}[a]{.48\textwidth}
  \centering
  \includegraphics[width=0.85\linewidth]{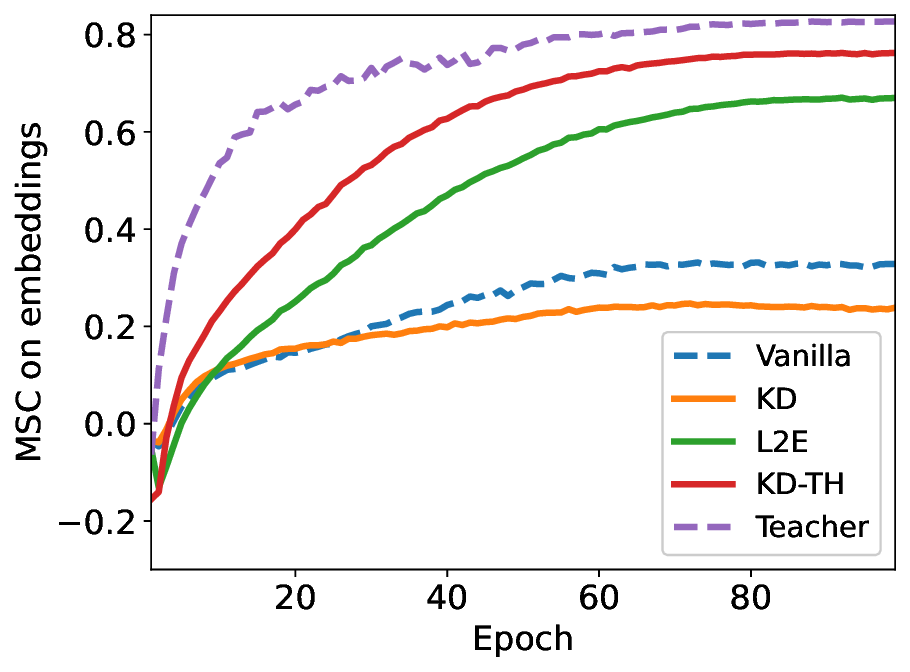}
  \caption{Stanford-cars}
\end{subfigure}%
\begin{subfigure}[h]{.48\textwidth }
  \centering
  \includegraphics[width=0.85\linewidth]{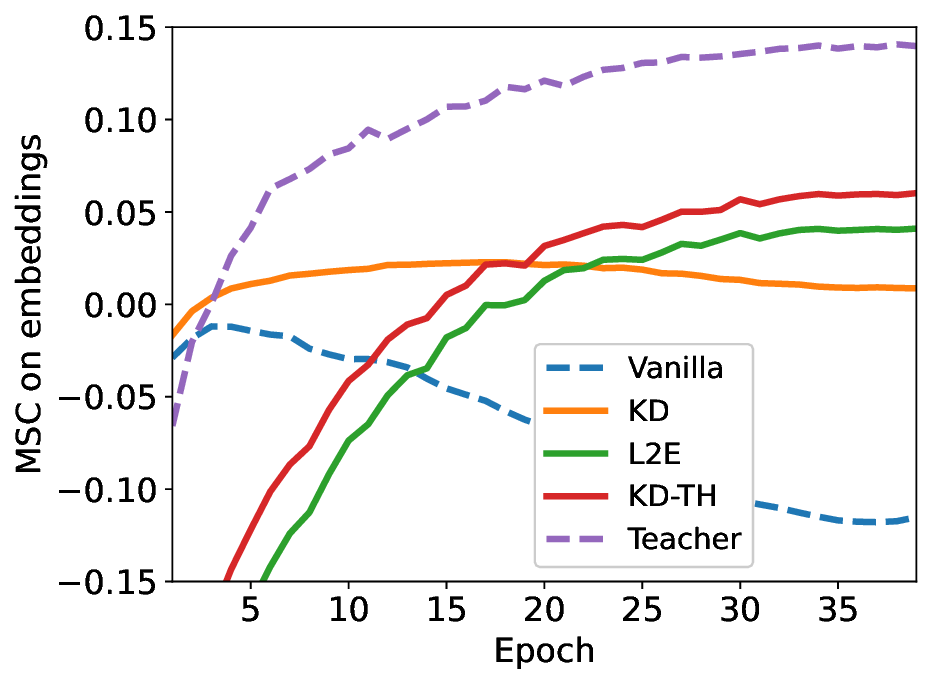}
  \caption{FoodX-251}
\end{subfigure}
\caption{\textbf{MSC score over training epochs for different KD methods.} TH-KD consistently increases the MSC score over the L2E baseline. Regular KD produces poor MSC score.}
\label{fig:MSC}
\vspace{-5mm}
\end{figure}

\subsection{Analysis} \label{sec: analysis}
In this section, we study notable properties of the proposed approaches, TH-KD and SH-KD, including their ability to accelerate the training process and how they impact the representation learning of the student.
\vspace{-3mm}

\subsubsection{Representation Similarity Between Teacher and Student.} \label{subsec:rep}
To demonstrate the ability to reliably transfer the representation knowledge from the teacher to the student, we measure the angle between corresponding embedding vectors extracted by the teacher and the student.
For each image the angle is given by,
\begin{equation} \label{eq:angles}
    d(\mathbf{z}_t, \mathbf{z}_s) = \text{arccos}\bigg(\frac{\mathbf{z}_t \cdot \mathbf{z}_s}{\|\mathbf{z}_t\|_2 \|\mathbf{z}_s\|_2 }\bigg).
\end{equation}
In Fig.~\ref{fig:angle}, we show the averaged angle between the teacher and student embeddings for Standford-cars, FoodX-251 and the face IJB-C datasets. The reported angle is obtained by averaging the instance angles $d(\mathbf{z}_t, \mathbf{z}_t)$ computed for all the test samples. 

Examination of Fig.~\ref{fig:angle} yields several observations.
First, the angle obtained by the regular KD is similar to the angle obtained in the vanilla model (between two unrelated models). This may arise from the fact that regular KD operates solely on the final network predictions. Therefore, KD does not offer any ability to transfer representational knowledge from the teacher to the student.
Second, L2E significantly improves both the similarity between the embedding vectors of the teacher and the student and the accuracy of the student. 
Third, the TH-KD and SH-KD further improve the similarity between the embedding vectors of the student and the teacher, as well as the classification accuracy. Fourth, using SH-KD considerably reduces the angle between the embedding vectors of the teacher and student models, across all datasets.

\begin{figure*}[t!]
\begin{subfigure}[a]{0.5\textwidth}\label{subfig:converge_cars}
  \centering
  \includegraphics[width=0.9\linewidth]{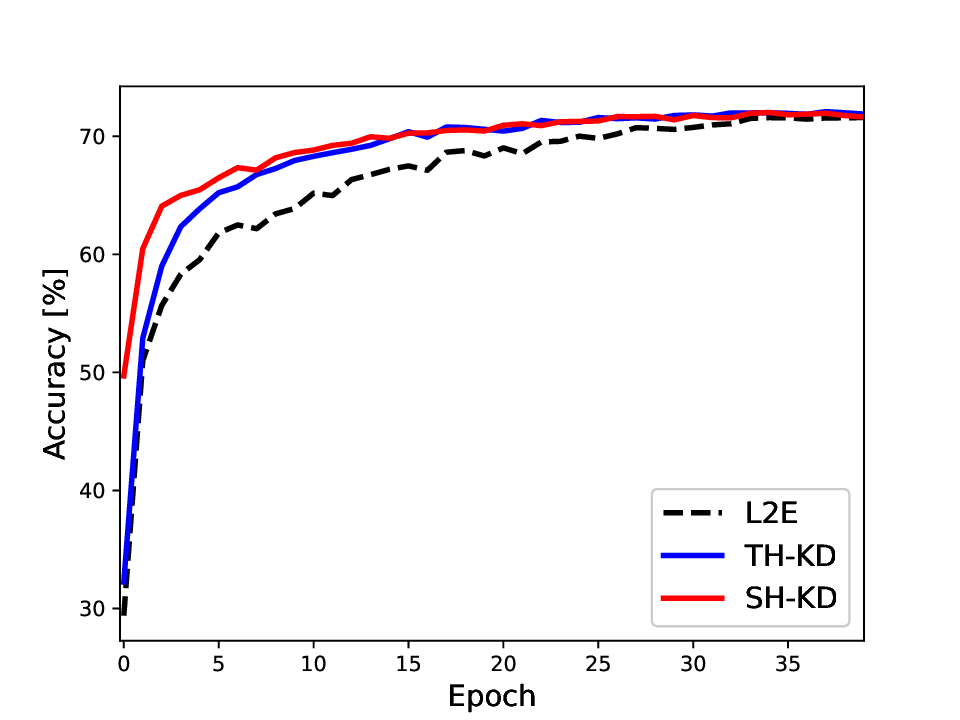}
  \caption{FoodX-251 dataset}
\end{subfigure}%
\begin{subfigure}[h]{0.5\textwidth }\label{subfig:converge_food}
  \centering
  \includegraphics[width=0.9\linewidth]{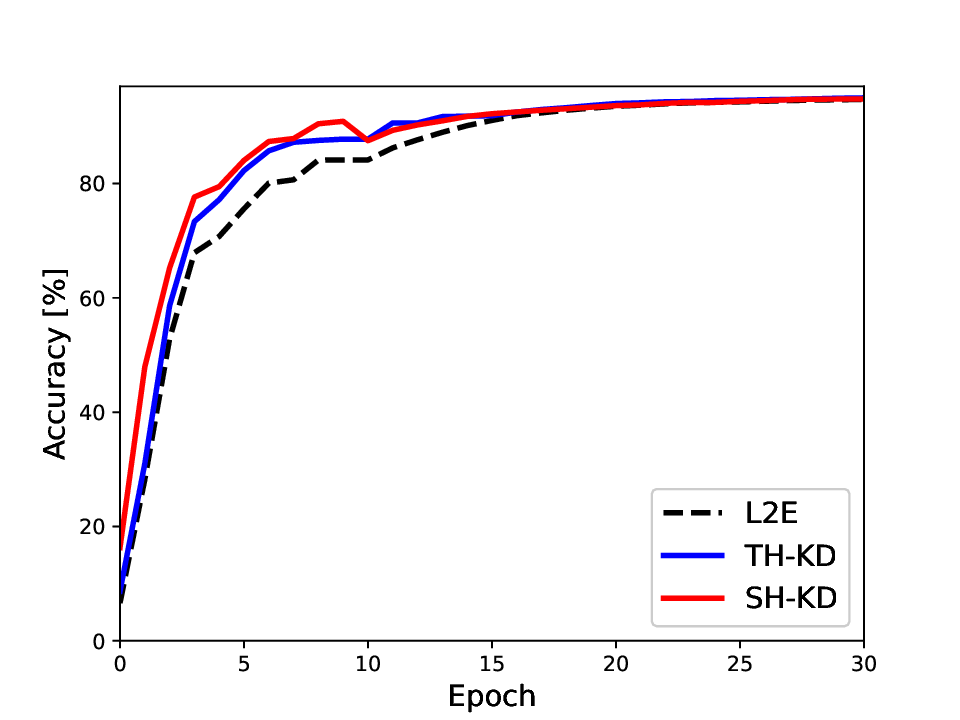}
  \caption{Stanford-cars dataset}
\end{subfigure}
\vspace{-3mm}
  \caption{\textbf{Training convergence.} 
  Training with TH-KD and SH-KD converges faster compared to regular training with L2E loss.}
\label{fig:accelerate}
\vspace{-3mm}
\end{figure*}

\vspace{-3mm}
\subsubsection{TH-KD Improves Representation Quality.} 
We use the mean silhouette coefficient (MSC) score \cite{silhouettes} to measure the clustering quality of the embeddings generated by the models' backbones, in terms of intra-class variation and inter-class separability. 
Let $\mathcal{Z}$ be the set of extracted embedding vectors: $\mathcal{Z}=\{\mathbf{z}_1, ..., \mathbf{z}_N \}$, where $N$ is the number of samples. The MSC is defined as,
\begin{equation}
    \text{MSC} = \frac{1}{N}\sum_{\mathbf{z} \in \mathcal{Z}}\frac{\eta(\mathbf{z}) - \sigma(\mathbf{z})}{\max\big(\eta(\mathbf{z}), \sigma(\mathbf{z})\big)},
\end{equation}
where $\sigma(\mathbf{z})$ is the averaged distance between $\mathbf{z}$ to the other embedding vectors residing in the same category as $\mathbf{z}$, and $\eta(\mathbf{z})$ is the minimum distance between $\mathbf{z}$ and the centers of the other categories. Higher $\eta(\mathbf{z})$ implies larger inter-class separability, and lower $\sigma(\mathbf{z})$ implies smaller intra-class variation. Typically, stronger models with higher capacity produce an embedding space with a higher MSC.

In Fig.~\ref{fig:MSC},  we plot the MSC score computed throughout the training epochs for several training modes: vanilla model trained without KD, KD, L2E, and TH-KD.
In both datasets (Stanford-cars and FoodX-251), training with TH-KD increases the MSC score compared to training with L2E. 
This supports the claim that training the student with the teacher head enables better representational knowledge transfer from the teacher to the student.
Note that in both datasets, the regular KD produces a poor MSC score. In the case of Stanford-cars the MSC is even degraded compared to the vanilla model.

\vspace{-3mm}
\subsubsection{KD Training Convergence.}\label{sec:convergence}
Often, training with knowledge distillation requires many more epochs than regular training \cite{beyer2021knowledge}. This is particularly true for representation distillation where the optimization involves the minimization of a distance function between the teacher and student embeddings. 
A slow training process increases the computational cost and limits resource utilization.   

In addition to improving a model's accuracy, we observe that networks trained using TH-KD and SH-KD converge faster compared to a baseline training with representation distillation. Classifier sharing reduces the number of trainable parameters and eases the training. TH-KD circumvents the learning of the student's classifier weights by initializing them with the teacher's head weights and freezing them. Moreover, SH-KD further accelerates the training because the features extracted by the teacher are more applicable for the student.
In Fig.~\ref{fig:accelerate}, we show the test accuracy on FoodX-251 and Stanford-cars, over the training epochs for TH-KD and SH-KD methods and compare them to a baseline training using the L2E loss. We used the TResNet-L as a teacher and OFA-62 as a student. As can be seen, both TH-KD and SH-KD result in faster training convergence compared to a baseline training with the L2E, and SH-KD further accelerates the training process.
To some extent, this compensates for the fact that SH-KD scheme requires an additional phase of training the initial student.

\begin{figure}[t!]
\begin{subfigure}[a]{0.33\textwidth}
  \centering
  \includegraphics[width=1\linewidth]{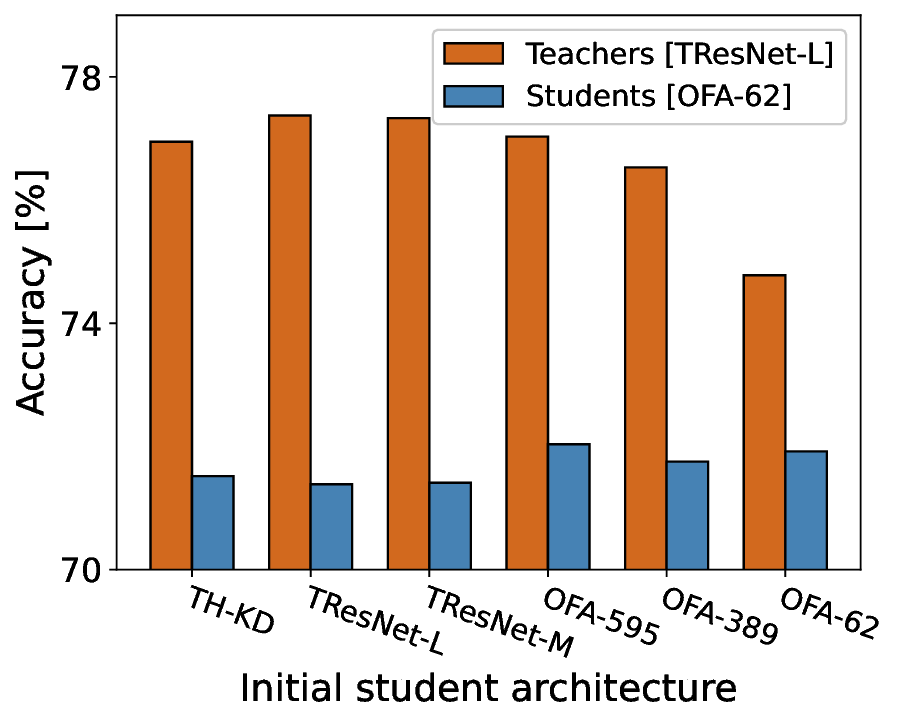}
  \caption{Accuracy}
  \label{subfig:shc_acc}
\end{subfigure}\hfill 
\begin{subfigure}[h]{0.33\textwidth}
  \centering
  \includegraphics[width=1\linewidth]{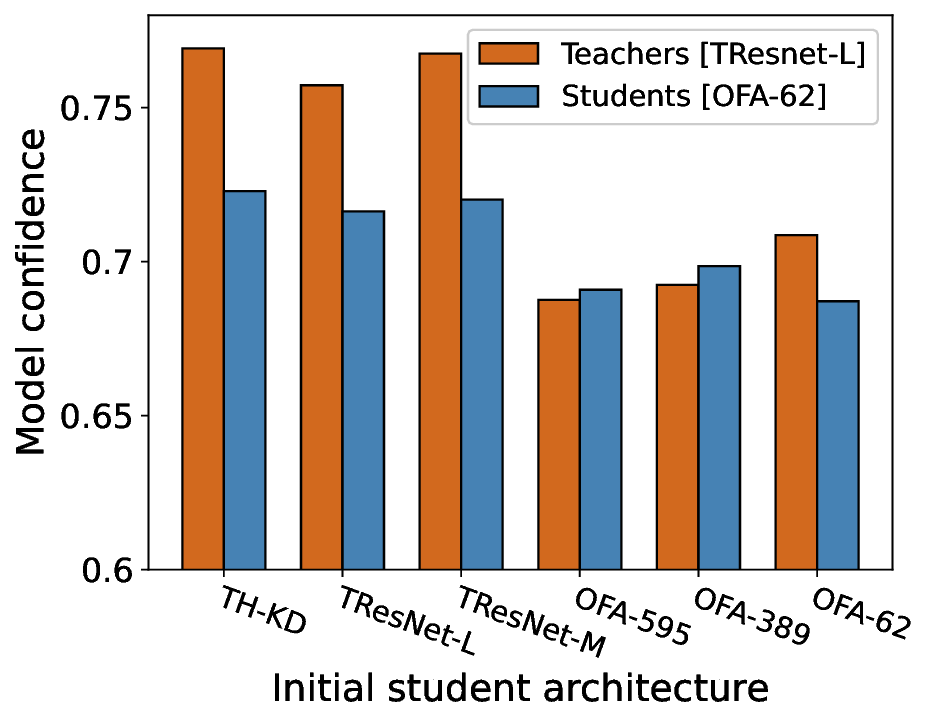}
  \caption{Confidence gap}
  \label{subfig:shc_conf}
\end{subfigure}\hfill 
\begin{subfigure}[h]{0.33\textwidth}
  \centering
  \includegraphics[width=1\linewidth]{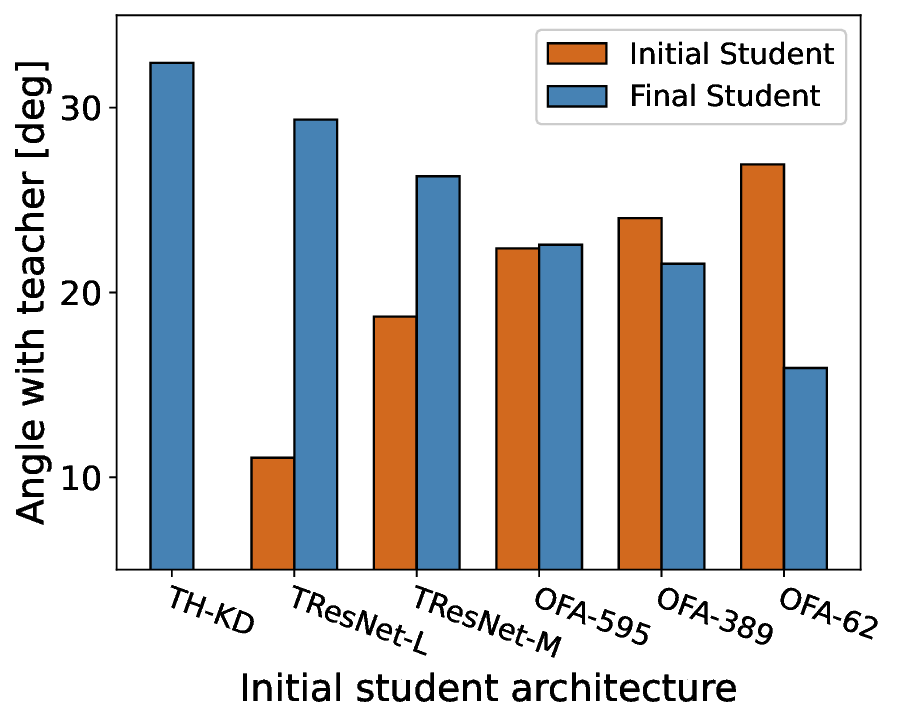}
  \caption{Embedding angle}
  \label{subfig:shc_angle}
\end{subfigure}
\caption{\textbf{Effect of initial student capacity on SH-KD.} 
We trained a set of teachers (TResNet-L), each with a head from an initial student of different capacity.  Then, we trained a final student (OFA-62) based on each teacher.
For each teacher-student pair, we report their accuracies, confidence gaps, and embedding angles on FoodX-251.}
\label{fig:shc_ablation}
\vspace{-5mm}
\end{figure}

\subsubsection{Effect of Initial Student Capacity on SH-KD.} 
In the SH-KD scheme, the teacher classifier head is replaced by the classifier head of the initial student.
Typically, a student model has lower capacity than the teacher. How does this capacity gap affect the trained teacher and the training of its final student?

We investigated how the capacity of the initial student affects the SH-KD training process, by examining a variety of initial student backbones. We used heads of students with different architectures to train different teacher models with similar backbones (TResNet-L). Then, we used each teacher to train a final student model (OFA-62). In Fig.~\ref{fig:shc_ablation}, we show the effects of the initial model selection on the outcome and process of SH-KD training. 

In Fig.~\ref{subfig:shc_acc} we report the accuracy of each teacher and its final student. We observe that decreasing initial student capacity reduces the teacher's accuracy, but may positively affect the final student's accuracy. Notably, using an initial student model from the OFA family yields a better final student than either using TH-KD or other architectures, even if the initial student architecture does not exactly match the final OFA-62 student architecture.

For the same settings, we present in Fig.~\ref{subfig:shc_conf} the confidence gap between the teacher and the student. Following \cite{xu2020computationefficient}, we measure the model confidence by the mean difference between first and second prediction values. A high confidence gap between the teacher and student predictions may imply a high capacity gap between the models \cite{guo2021reducing}.
As can be seen, the confidence gap between the teacher and the final student (whose architectures are fixed) is substantially reduced when the initial student model belongs to the OFA family. This is an indication that SH-KD can mitigate the capacity gap between student and teacher by adapting the properties of the teacher to match the capacity of the student.

To further understand the SH-KD process and its outcomes, we present the difference between teacher's and student's embedding vectors, as measured by their angle (see equation (\ref{eq:angles})), in Fig.~\ref{subfig:shc_angle}. On one hand, the difference between the embedding vectors of each initial student model and its matching teacher increases accordingly when their capacity gap increases. On the other hand, for the final student, its embedding vectors match the teacher's embedding vectors more closely if its teacher was trained with a smaller initial student model. Thus, using an initial student of a capacity matching the low capacity of the final student constraints the trained teacher, and therefore allows the final student to match it better, and eventually yield better performance.

\section{Conclusion}
\label{conclusion}
\vspace{-3mm}
In this paper we explored two techniques for representation distillation that are based on classifier sharing between the teacher and the student. The TH-KD approach shares the teacher's classifier with the student to constrain the representation distillation process. The SH-KD enables sharing the student's classifier within the teacher's training at the cost of another training iteration. 
Extensive experiments and analyses demonstrate the effectiveness of the proposed schemes on various domains and datasets.
We show that both TH-KD and SH-KD accelerate the representation distillation process. Moreover, the TH-KD technique is useful in improving the discrimination power of the embeddings extracted by the student's backbone. Finally, training with SH-KD and TH-KD increases the similarity between the teacher and student embeddings and leads to the desired improvement in the student accuracy.
For future work, we would like to investigate ways to apply the proposed approaches to an ensemble of teachers. In addition, a theoretical formulation based on tighter upper bounds may yield a better understanding of the possible benefits and limitations of the proposed methods.


%
%
\bibliographystyle{splncs04}
\bibliography{egbib}

\begin{appendices}

\section{Training Details} \label{appendix:methods}
\subsection{CIFAR-100 Training Details}
For a fair comparison, we used the public code provided for the CRD work \cite{tian2020contrastive} and followed the same experimental protocol.

\subsection{Fine-grained Classification Training Details}
We tested our approaches on two fine-grained classification datasets; FoodX-251 \cite{foodx_dataset_2019} and Stanford Cars \cite{stanford_cars_2013}. We used both modern network architectures of TResNet \cite{ridnik2021imagenet21k} and OFA \cite{ofa_2020} and classical ResNet architectures \cite{resnet_2015}. Specifically, we tested six training configurations with teachers: TResNet-L, TResNet-M, and ResNet101, and students: OFA-62, OFA-389, OFA-595, ResNet18 and ResNet26.

We trained all models with a combination of a base cross-entropy loss, a triplet-loss \cite{2015}, and the specified KD losses. The Stanford-cars dataset was trained for 100 epochs with a learning-rate of 5e-4 and a weight decay of 2e-4. The FoodX-251 dataset was trained for 40 epochs with a learning-rate of 3e-4 and a weight decay of 1e-4. In all experiments, we used the Adam optimizer \cite{kingma2017adam} with a cosine decay learning-rate schedule. All models were pre-trained on the ImageNet-21k dataset \cite{ridnik2021imagenet21k}. The input image size was 224 x 224. The embedding dimension was 2,048. For regularization, we used standard augmentation techniques \cite{cubuk2019autoaugment}. We used a single V100 machine for seach run.  
For TH-KD and SH-KD, we used $\alpha^{\text{TH}}=1$ in equation (\ref{TH_CE_loss}) of the paper.
We used L2E loss with $\beta=0.05$.

\subsection{Training Details for Face Verification}
Our implementation is based on the PyTorch framework and combines different blocks from the repositories Insightface \cite{deng2018arcface} and timm \cite{rw2019timm}. We used the same hyper-parameters for each training process.

The models were trained for 30 epochs, with initial learning-rate of 1e-2 and cosine decay schedule. We set the weight decay to 1e-4, except for the classification layer which did not have weight decay at all.
We used the RMSprop optimizer with momentum of 0.9.
The input faces were normalized into a patch of size 112 x 112, using the alignment method from \cite{deng2018arcface}. We used 3 types of data augmentations techniques : random horizontal flip (probability of 0.5), color jitter (brightness, contrast and saturation jitter of ${\pm0.4}$), random erasing (probability of 0.1) \cite{zhong2020random}. 
For the CosFace loss we used the constants ${s=64}$ and ${m=0.4}$.
We used the L2E loss with $\alpha=0$ and $\beta=5$. The TH-KD scheme was performed with $\alpha^{\text{TH}}=1$ in equation (\ref{TH_CE_loss}) of the paper. The embedding dimension used in all methods was 512.

\end{appendices}

\end{document}